\documentclass{article} 
\usepackage{iclr2027_conference,times}

\usepackage{amsmath,amsfonts,bm}

\def\eqref#1{equation~\ref{#1}}

\def\1{\bm{1}}

\DeclareMathAlphabet{\mathsfit}{\encodingdefault}{\sfdefault}{m}{sl}
\SetMathAlphabet{\mathsfit}{bold}{\encodingdefault}{\sfdefault}{bx}{n}

\usepackage{hyperref}
\usepackage{url}
\usepackage{graphicx}
\usepackage{amsmath}
\usepackage{amssymb}
\usepackage{bm}
\usepackage{booktabs}
\usepackage{makecell}
\usepackage{multirow}
\usepackage{algorithm}
\usepackage{algorithmic}

\title{LazyFlash: Forward-Dense, Backward-Sparse Attention for Accelerated Long-Context Fine-Tuning}

\author{Keqi Deng, Ruchao Fan, Jinyu Li\\
Microsoft, USA \\
\texttt{keqideng@microsoft.com} \\
}

\iclrfinalcopy 
\begin{document}

\maketitle
\vspace{-0.3cm}
\begin{abstract}

As large language models (LLMs) process increasingly long sequences, attention becomes a major computational bottleneck. Sparse attention mostly targets inference, yet many methods still train densely or cannot support sparse training, and their added complexity makes sparsity hard to turn into wall-clock speedups, leaving dense attention widely used in practice. This paper introduces LazyFlash, a forward-dense, backward-sparse attention method built on the observation that attention training is dominated by the backward pass. LazyFlash keeps the forward pass exactly dense, so the trained model remains a standard dense model that existing systems such as vLLM can serve without any change. For the backward pass, LazyFlash proposes a gradient-aware block sparsification design that ranks query blocks by their attention-output gradient norms and skips those contributing negligibly to parameter updates, turning the dense backward into a block-sparse one. A gradient-mass budget adapts the amount of skipped computation to each head and training step, while importance sampling preserves the dense gradient in expectation. These designs are implemented inside the tiled FlashAttention backward kernel, where skipped blocks are never loaded or computed, translating gradient sparsity into real wall-clock acceleration. Experiments on long-context text and speech LLMs show that LazyFlash maintains downstream performance while achieving up to 5.7× attention-backward speedup over FlashAttention and 1.7× end-to-end training speedup.

\end{abstract}

\section{Introduction}

Large language models (LLMs) have rapidly expanded their context windows from a few thousand to hundreds of thousands of tokens \citep{achiam2023gpt, dubey2024llama, comanici2025gemini}, and self-attention remains the central component behind this capability. Despite a long line of efficient alternatives, standard full attention is still the dominant choice in mainstream LLMs, since it preserves exact token interactions and is backed by highly optimized kernels. Its cost, however, grows quadratically with sequence length, which makes attention the dominant cost of long-context training: doubling the context quadruples the leading attention
computation. 
FlashAttention and its successors \citep{dao2022flashattention, dao2023flashattention2, shah2024flashattention3} have largely removed the memory bottleneck by tiling the computation so that the full attention matrix is never
materialized in high-bandwidth memory (HBM), and they have become the standard attention implementation for LLM training.
These kernels, however, reduce memory traffic rather than arithmetic, so the quadratic FLOPs remain and further gains from kernel-level engineering are increasingly incremental. The same pressure is not confined to text. Speech and audio LLMs increasingly train on hours-long recordings \citep{DBLP:conf/icml/ParkSJKRS25, peng2025vibevoice, deng2026unique}, where a single utterance already spans tens of thousands of positions. Accelerating long-context attention training is therefore a shared challenge across modalities rather than a text-only concern.

Sparse attention is a natural candidate for reducing this cost, but most existing methods target inference rather than training. Training-free approaches \citep{DBLP:conf/nips/Zhang00CZC0TRBW23, DBLP:conf/icml/TangZZXKH24, DBLP:conf/nips/XiaoZ0XLZ0024} select a small subset of the KV cache at decoding time and leave training entirely dense, offering no training-side savings. Trainable sparse
attention instead incorporates sparsity into the forward pass so that the model adapts to sparse decoding. Native Sparse Attention (NSA) \citep{DBLP:conf/acl/YuanGD0ZZXWW0WR25} is designed for pre-training from scratch and departs substantially from standard attention, so its gains do not carry over to models that keep standard attention. SeerAttention \citep{gao2025seerattention} and DeepSeek Sparse Attention (DSA) \citep{liu2025deepseek} learn their sparse selection by distilling from the dense attention distribution, which requires dense attention to be computed as the target during training and therefore precludes any training-time saving. Mixture of Block Attention (MoBA) \citep{lu2025moba} comes closer to a trainable accelerator, but its block selection introduces extra computation on top of the sparse attention itself, and decoding the resulting model densely creates a train-inference mismatch that can degrade quality. More broadly, the additional importance estimation, top-$k$ selection, and irregular sparse kernels mean that theoretical FLOP reductions do not always translate into wall-clock gains unless the context is extremely long, which is why dense attention remains widely used in deployment. 
Although sparsity is well aligned with the intrinsic structure of attention, how to
exploit it to accelerate the training of standard attention, rather than its inference,
remains an open challenge.

This paper proposes LazyFlash, a backward-sparse attention method that accelerates the
training of standard attention. Attention is sparse in both its forward and its backward
computation, but prior work has concentrated almost entirely on the forward side, even
though the backward pass accounts for most of the training cost. LazyFlash therefore
leaves the forward computation dense and exact and introduces sparsity only in the
backward pass, implemented on top of FlashAttention. Since the forward pass is untouched,
the trained model remains an ordinary dense-attention model and can be served directly by
mature inference frameworks such as vLLM \citep{DBLP:conf/sosp/KwonLZ0ZY0ZS23}, which is
of practical value because no change to the deployment pipeline is required. LazyFlash
operates at the granularity of query blocks aligned with the FlashAttention \citep{dao2023flashattention2} tiling, and
measures the contribution of each block to the attention gradient from the norm of its
output gradient. Blocks whose contribution is negligible are dropped, so their tiles are
never loaded or computed in the backward kernel. Rather than fixing the skip ratio in
advance, LazyFlash sets a budget on the fraction of gradient mass that may be discarded,
so that the resulting sparsity adapts to how concentrated the gradients are at each step.
Within this budget, blocks are kept by sampling with probability proportional to their
gradient contribution and are rescaled accordingly, which keeps the estimated gradient unbiased. Experiments on long-context text and long-form speech LLMs show that LazyFlash matches the quality of dense training while reducing the attention backward time by up to $5.7\times$ and the overall training time by up to $1.7\times$.

The main contributions of this paper can be summarized in four parts:
\begin{itemize}
    \item LazyFlash, a forward-dense and backward-sparse attention method that speeds up
    long-context fine-tuning while keeping the forward pass exact, so the trained model
    remains a standard dense model that needs no change at inference.

    \item A block selection method that scores each query block by the norm of its output
    gradient and skips the blocks that barely contribute, with a gradient-mass budget
    that sets the skip ratio from the gradient distribution rather than from a fixed
    threshold.

    \item An importance sampling scheme that keeps blocks with probability proportional
    to their gradient contribution and rescales them accordingly, so the gradient
    estimate stays unbiased instead of repeatedly dropping the same blocks.

    \item A FlashAttention backward kernel supporting both dense and variable-length
    inputs, reaching up to $5.7\times$ faster attention backward and $1.7\times$ faster
    end-to-end training on text and speech LLMs.
\end{itemize}

\section{Related Work}

\subsection{Efficient Attention Kernels}

FlashAttention \citep{dao2022flashattention} tiles the computation and uses an online
softmax so that the full attention matrix is never written to HBM, FlashAttention-2
\citep{dao2023flashattention2} improves the work partitioning across thread blocks, and
FlashAttention-3 \citep{shah2024flashattention3} exploits asynchrony and low precision on
newer hardware. For inference, PagedAttention \citep{DBLP:conf/sosp/KwonLZ0ZY0ZS23}
organizes the KV cache into pages, and FlashInfer \citep{ye2025flashinfer} provides
customizable decoding kernels for batched serving. These kernels all compute exact
attention and cut memory traffic, but the arithmetic still grows quadratically with
sequence length. LazyFlash builds on this line of work and is complementary to it, since
it removes backward computation rather than moving the same computation more efficiently.

\subsection{Sparse Attention}

Sparse attention computes only part of the attention map. Training-free methods exploit
the sparsity already present in dense-trained models: H2O
\citep{DBLP:conf/nips/Zhang00CZC0TRBW23} evicts KV entries with low accumulated attention
scores, Quest \citep{DBLP:conf/icml/TangZZXKH24} selects the top-$k$ KV pages per query
from min-max bounds, InfLLM \citep{DBLP:conf/nips/XiaoZ0XLZ0024} extends page selection to
million-token contexts, SpargeAttention \citep{DBLP:conf/icml/ZhangXHWX0C25} and BLASST
\citep{yuan2025blasst} skip computation by softmax thresholding, and MInference
\citep{jiang2024minference} and FlexPrefill \citep{DBLP:conf/iclr/LaiLLMZ25} target the
prefilling stage. Trainable methods instead place the sparse pattern in the forward pass
so that the model adapts to it: NSA \citep{DBLP:conf/acl/YuanGD0ZZXWW0WR25} combines
compressed, selected and sliding branches and is trained from scratch, MoBA
\citep{lu2025moba} routes each query to a few blocks, SeerAttention
\citep{gao2025seerattention} trains a gate by self-distillation against the dense attention
map, and DSA \citep{liu2025deepseek} learns a separate indexer to score the cache. 
Neither family, however, aims at accelerating the training of standard attention:
training-free methods act only at decoding time, and trainable ones change the forward
pass and therefore the model itself. Sparsifying the backward pass has been studied
outside attention, where meProp \citep{sun2017meprop} keeps only the top-$k$ entries of
the output gradient of each linear layer, but such element-wise deterministic truncation
is defined for linear layers and does not transfer to attention, whose backward pass is
computed in tiles and calls for a design of its own. How to make the backward pass of attention sparse, while keeping the forward pass exact,
is therefore still challenging.

\section{Methodology}
\label{sec:method}

\begin{figure*}[t]
\centering
\includegraphics[width=\textwidth]{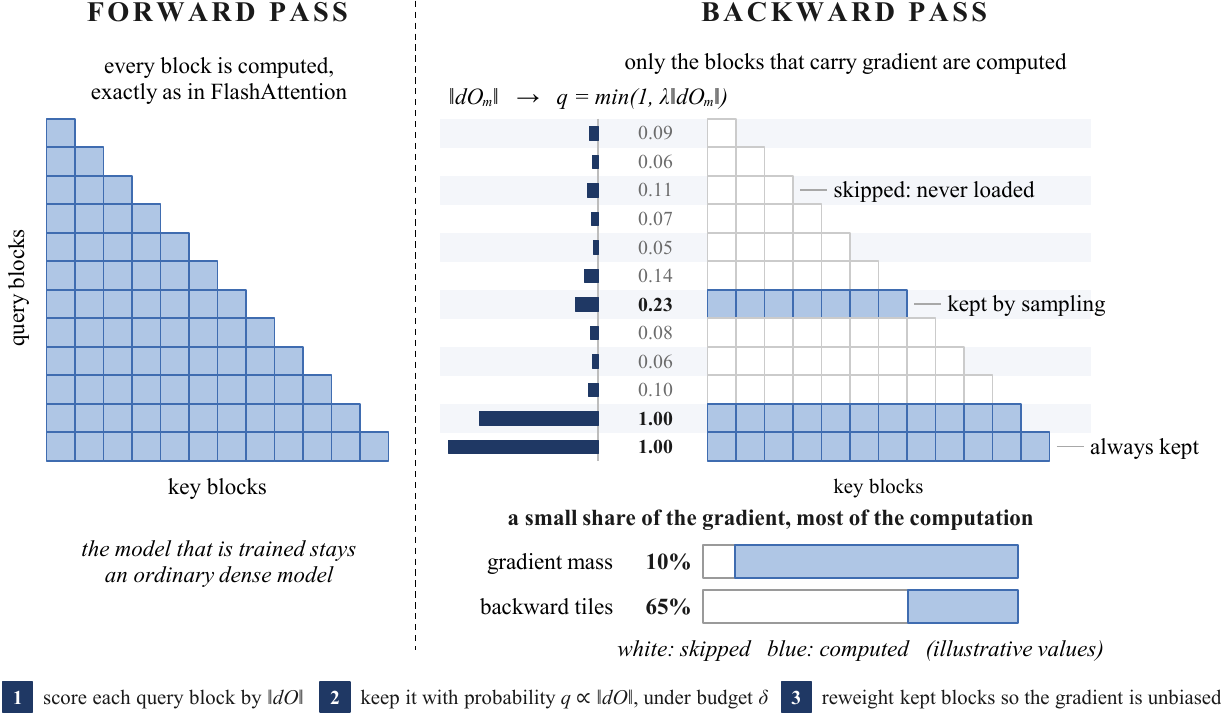}
\caption{Overview of LazyFlash. The forward pass computes every block, so the trained model remains an ordinary dense-attention model. In the backward pass, each query block is kept with a probability set by its output-gradient norm: skipped tiles are never loaded, and kept blocks are reweighted so that the gradient stays unbiased. As the bars illustrate, a small budget on gradient mass translates into a much larger saving in computation.}
\label{fig:overview}
\end{figure*}

This section presents LazyFlash, which accelerates attention training by computing the backward pass only where it matters while leaving the forward pass exact. Two observations motivate the design. 
The backward pass performs about twice the work of the forward pass, and the gradient it carries is spread very unevenly across query blocks, so a block that receives little output gradient contributes little to any parameter gradient. 
The signal is inexpensive to obtain because the attention-output gradient is already available when the backward pass begins. Scoring a block requires only a norm over this existing tensor, whereas typical sparse forward attention must estimate importance before it can compute anything.
LazyFlash makes two separate decisions. The amount of computation that may be removed is set by a budget $\delta$ on the share of gradient mass that can be discarded, so the sparsity adapts to the gradient distribution of each head at each step (Section~\ref{sec:selection}). 
Which blocks are kept is then decided by sampling in proportion to their gradient, with a compensating rescaling that leaves the gradient unbiased (Section~\ref{sec:sampling}). 
Both steps run in a compact CUDA pipeline that produces per-block weights for
the FlashAttention backward kernel. Tiles with zero weight are skipped before
their data are loaded (Section~\ref{sec:kernel}).
Figure~\ref{fig:overview} contrasts the forward and backward passes.

\subsection{The Attention Backward Pass}
\label{sec:prelim}

Let $\mathbf{Q},\mathbf{K},\mathbf{V}\in\mathbb{R}^{n\times d}$ be the queries, keys, and values of one head, with $n$ the sequence length and $d$ the head dimension. The forward pass computes the attention scores, the attention weights, and the output:
\begin{equation}
\mathbf{S}=\frac{\mathbf{Q}\mathbf{K}^{\top}}{\sqrt{d}},
\qquad
\mathbf{P}=\mathrm{softmax}(\mathbf{S}),
\qquad
\mathbf{O}=\mathbf{P}\mathbf{V}.
\end{equation}
Given the output gradient $\mathrm{d}\mathbf{O}$, the backward pass computes
\begin{equation}
\mathrm{d}\mathbf{V}=\mathbf{P}^{\top}\mathrm{d}\mathbf{O},
\quad
\mathrm{d}\mathbf{P}=\mathrm{d}\mathbf{O}\,\mathbf{V}^{\top},
\quad
\mathrm{d}\mathbf{Q}=\frac{\mathrm{d}\mathbf{S}\,\mathbf{K}}{\sqrt{d}},
\quad
\mathrm{d}\mathbf{K}=\frac{\mathrm{d}\mathbf{S}^{\top}\mathbf{Q}}{\sqrt{d}},
\label{eq:bwd}
\end{equation}
where $\mathrm{d}\mathbf{S}$ is obtained from $\mathrm{d}\mathbf{P}$ through the row-wise softmax Jacobian, requiring only $\mathcal{O}(n^2)$ operations. The forward pass therefore involves two matrix products of order $n^{2}d$ and the backward pass involves four. Consequently, at long context where $n\gg d$, the backward pass performs roughly twice the work of the forward pass and dominates the computation during training.

Two properties of Eq.~\ref{eq:bwd} are used below. First, FlashAttention evaluates it tile by tile over a grid of query and key blocks, so computation can be removed at the granularity of a block rather than of an individual entry. 
Second, all four products are driven by $\mathrm{d}\mathbf{O}$: it enters
$\mathrm{d}\mathbf{V}$ and $\mathrm{d}\mathbf{P}$ directly, $\mathrm{d}\mathbf{P}$ is turned
into $\mathrm{d}\mathbf{S}$ by the softmax Jacobian, and $\mathrm{d}\mathbf{S}$ in turn
produces $\mathrm{d}\mathbf{Q}$ and $\mathrm{d}\mathbf{K}$.
Each row block of $\mathrm{d}\mathbf{Q}$ comes from the matching row block of $\mathrm{d}\mathbf{O}$, whereas $\mathrm{d}\mathbf{K}$ and $\mathrm{d}\mathbf{V}$ accumulate contributions from every query row. A query block whose output gradient is small therefore contributes little to $\mathrm{d}\mathbf{Q}$, $\mathrm{d}\mathbf{K}$ or
$\mathrm{d}\mathbf{V}$, and its tiles are natural candidates for removal.

\subsection{Gradient-Aware Block Selection}
\label{sec:selection}

The sequence is divided into $M=\lceil n/B\rceil$ query blocks of $B$ rows, matching the tile size of the backward kernel, and $\mathrm{d}\mathbf{O}_m$ denotes the output gradient of the $m$-th block. Each block receives a single score
\begin{equation}
c_m=\|\mathrm{d}\mathbf{O}_m\|_F ,
\end{equation}
which measures how much gradient enters that block. Computing it requires no extra pass over the data, since $\mathrm{d}\mathbf{O}$ is already available when the backward pass begins.

A fixed skip ratio is a poor way to turn these scores into a decision, because the
concentration of $c_m$ varies from step to step and from layer to layer: the same ratio discards useful gradient in one step and leaves easy savings unused in another. LazyFlash instead constrains the share of gradient mass that may be discarded. Writing $c_{(1)}\le\cdots\le c_{(M)}$ for the scores of one head in ascending order, the number of
blocks that may be dropped is
\begin{equation}
k^{\star}=\max\Big\{\,k\ :\ \sum_{i=1}^{k}c_{(i)}^{2}\ \le\ \delta\sum_{m=1}^{M}c_m^{2}\,\Big\},
\label{eq:budget}
\end{equation}
so that the discarded blocks together hold at most a fraction $\delta$ of the gradient
energy of that head. Note that $\delta$ is a relative quantity rather than a threshold on
$\|\mathrm{d}\mathbf{O}\|$, so the sparsity it produces adapts to the gradient distribution
of each head at each step. It is also the only quantity exposed to the user. 
Under supervised fine-tuning the loss covers only part of the sequence, so the scores $c_m$
are far from uniform. A small $\delta$ then allows a large $k^{\star}$, which is why a
budget expressed in gradient mass yields a much larger reduction in computation
(as shown in Figure~\ref{fig:overview}).

\subsection{Unbiased Importance Sampling}
\label{sec:sampling}

Dropping the $k^{\star}$ lowest-scoring blocks outright would bias training, because the
same blocks are excluded at every step and their contribution never reaches the parameter
gradients. LazyFlash therefore uses the budget to determine a target number of kept blocks
and realizes that target in expectation.
Block $m$ is kept with probability
\begin{equation}
q_m=\min(1,\ \lambda c_m),
\qquad\text{with }\lambda\text{ chosen so that }\sum_{m=1}^{M}q_m=M-k^{\star},
\label{eq:qm}
\end{equation}
where $\lambda$ is found by bisection. Writing $z_m\sim\mathrm{Bernoulli}(q_m)$ for the
outcome of that draw, and $\mathbf{g}_m$ for the contribution of block $m$ to a parameter
gradient $\mathbf{g}=\sum_m\mathbf{g}_m$, the backward pass evaluates
\begin{equation}
\hat{\mathbf{g}}=\sum_{m=1}^{M}\frac{z_m}{q_m}\,\mathbf{g}_m .
\label{eq:estimator}
\end{equation}
Since $\mathbb{E}[z_m]=q_m$, every term has expectation $\mathbf{g}_m$ and
$\mathbb{E}[\hat{\mathbf{g}}]=\mathbf{g}$ for any $\lambda>0$, so the estimate is unbiased.
The rescaling has a simple reading: a block that is evaluated once every $1/q_m$ steps
counts for $1/q_m$ blocks when it does appear, so its full contribution is recovered on
average.

Unbiasedness holds for any positive keep probability, so the role of Eq.~\ref{eq:qm} is not
to remove the bias but to keep the variance small. Sampling in proportion to the gradient
magnitude has two effects. Blocks with large $c_m$ saturate at $q_m=1$, so they are never
dropped and never rescaled, and the randomness is confined to the low-gradient tail. Within
that tail $\|\mathbf{g}_m\|$ is itself roughly proportional to $c_m$, so a kept contribution
has magnitude $\|\mathbf{g}_m\|/q_m\approx 1/\lambda$, bounded independently of how small
the block is. A uniform keep probability would instead leave the variance dominated by the
largest blocks. 
Sampling in proportion to gradient magnitude is a standard variance-reduction device in
stochastic optimization \citep{katharopoulos2018not}, where it is applied to training
examples; LazyFlash applies it to the blocks of an attention kernel.

\subsection{Kernel Realization}
\label{sec:kernel}

\textbf{CUDA selection pipeline.} Scoring, budgeting and sampling run in a compact CUDA
pipeline. It reads $\mathrm{d}\mathbf{O}$ and writes one weight for each entry of the block
grid, equal to $1/q_m$ when the block is kept and to zero when it is skipped. The backward
kernel reads these weights as it walks the grid. A zero weight makes it skip the tile before
any data is loaded, so memory traffic is saved as well as arithmetic, and a non-zero weight
scales the tile as it is accumulated. The forward kernel is untouched, so training produces
an ordinary dense-attention model that runs on a standard inference stack.

\textbf{Variable-length inputs.} LazyFlash supports the dense interface and the packed
variable-length one. In the packed case the weight grid is indexed within each sub-sequence,
so a sub-sequence has to begin at a multiple of $B$. LazyFlash pads each one to the next
multiple and removes the padding afterwards. Padded rows carry no gradient, so their score
is zero and they are skipped.

\textbf{Overhead.} The selector reads $\mathrm{d}\mathbf{O}$ once and produces $M$ scores per
head, a small cost next to the $O(n^{2})$ work that is removed.

\section{Experiments}
\subsection{Setup}
\label{sec:setup}

LazyFlash is evaluated on two modalities, long-context text understanding and long-form
automatic speech recognition (ASR). For text, Qwen3-8B-Base~\citep{yang2025qwen3} is
fine-tuned on an equal-proportion mixture of
LongAlign-10k~\citep{DBLP:conf/emnlp/BaiLZHQH0DL24}, a long-context instruction corpus of
about 10K instances, and general instruction data from
ShareGPT,\footnote{The cleaned split released as
\texttt{ShareGPT\_V3\_unfiltered\_cleaned\_split}.} so that long-context adaptation does not
come at the expense of general instruction following. The fine-tuned models are evaluated on
LongBench-Pro~\citep{chen2026longbench}, a bilingual benchmark of 1{,}500 English and
Chinese samples covering 11 primary and 25 secondary tasks with inputs up to 256K tokens.
For speech, 50K hours of English long-form data, with single utterances lasting up to half
an hour, are used to train the speech LLM, and evaluation is performed on English long-form
ASR test sets of 10-minute utterances.

Qwen3-8B-Base is used as the text LLM. The speech LLM uses the same language backbone,
Qwen3-8B~\citep{yang2025qwen3}, coupled with a Conformer speech encoder, so that the two
settings share one attention implementation while differing in input modality and in
sequence statistics.

Dense attention, as implemented by FlashAttention-2~\citep{dao2023flashattention2}, is the
baseline that LazyFlash is compared against, and MoBA~\citep{lu2025moba} is implemented as a
representative sparse attention method and run on both modalities, with a block size of 512
and top-4 selection for text and top-3 for speech, which corresponds to a budget of 2048 and
1536 tokens per query respectively. All runs share the same
backbone, the same data and data order, and the same number of training steps, so that any
difference in quality or in wall-clock time comes from the attention backend alone. The
budget is set to $\delta=0.1$ for text and $\delta=0.4$ for speech unless otherwise stated.
Ablations in Section~\ref{sec:ablation} examine what makes the selection work, and include a top-$k$ rule that extends meProp~\citep{sun2017meprop} from linear layers to
attention blocks. More details refer to Appendix~\ref{app:hyper} and~\ref{app:repro}.

\subsection{Long-Context Text Understanding}
\label{sec:text}


\begin{table*}[t]
\centering
\small
\setlength{\tabcolsep}{3.5pt}
\begin{tabular}{l|c|cc|cccc|cc}
\toprule
\multirow{2}{*}{Method} & \multirow{2}{*}{Overall} & \multicolumn{2}{c|}{Language}
& \multicolumn{4}{c|}{Difficulty} & \multicolumn{2}{c}{Training attn.\ (ms)} \\
& & En & Zh & Extreme & Hard & Moderate & Easy & fwd & bwd \\
\midrule
Qwen3-8B-Base (no FT) & 19.70 & 19.35 & 19.78 & 19.39 & 16.89 & 13.98 & 24.67 & -- & -- \\
\midrule
Dense attention & 29.64 & 30.31 & 28.63 & 25.18 & 25.40 & 19.54 & \textbf{41.78} & 547.3 & 1429.8 \\
MoBA & & & & & & & & 700.3 & 710.2 \\
\quad sparse inference & 24.48 & 24.73 & 24.23 & 20.04 & 21.71 & 15.36 & 35.65 & & \\
\quad sparse prefill & 23.64 & 24.99 & 22.30 & 20.27 & 19.21 & 14.46 & 34.89 & & \\
\quad dense inference & 29.37 & 30.16 & 28.58 & \textbf{25.58} & 27.35 & 19.46 & 39.98 & & \\
\textbf{LazyFlash} & \textbf{30.24} & \textbf{30.37} & \textbf{29.03} & 25.22 & \textbf{27.51}
& \textbf{20.04} & 40.88 & 562.2 & \textbf{507.4} \\
\bottomrule
\end{tabular}
\caption{LongBench-Pro results after fine-tuning Qwen3-8B-Base (higher is better,
$\uparrow$), with the attention time per microbatch measured during training. The three
indented rows are one MoBA-trained model served in three ways: sparse attention at prefill
and decode, sparse attention at prefill only, and fully dense attention. They share the
training cost listed on the MoBA row.}
\label{tab:text}
\end{table*}

Table~\ref{tab:text} reports LongBench-Pro after fine-tuning Qwen3-8B-Base, together with
the attention time measured during training. Fine-tuning on the long-context mixture data raises
the overall score from 19.70 to 29.64, so this is a setting in which long-context training
matters and a degradation caused by the attention backend would be visible.

LazyFlash is the strongest of the fine-tuned models on the overall score, 30.24 against
29.64 for dense attention, and it leads on five of the seven columns, with the largest gain
on the Hard split (27.51 against 25.40). It is behind on Easy and marginally behind MoBA
with dense inference on Extreme, and neither changes the ordering on the aggregate score.
Removing most of the attention backward therefore leaves quality unaffected.


MoBA has to be evaluated under more than one inference configuration, because it is trained
with a sparse forward pass and the trained model inherits that choice. Used as designed it
reaches 24.48, and restricting the sparsity to the prefill stage does not help (23.64). It
approaches dense training only when served with fully dense attention (29.37), which gives
up the inference sparsity that motivated it. LazyFlash has a dense forward pass by
construction and therefore a single inference behaviour.

The timings show where the saving comes from. LazyFlash spends 507.4~ms in the backward pass
against 1429.8~ms for dense attention, a speedup of $2.82\times$, while its forward pass is
bit-exact and its cost is essentially unchanged, 562.2~ms against 547.3~ms. MoBA moves in
the opposite direction: its forward pass must build block representatives, score them
against every query and select the top-$k$ blocks before any attention is computed, which
raises the forward cost to 700.3~ms, $28\%$ above dense, while its backward pass at
710.2~ms is only $2.01\times$ faster than dense and $1.40\times$ slower than LazyFlash.
Section~\ref{sec:efficiency} examines how these gains scale with context length.

\subsection{Long-Form Speech Recognition}
\label{sec:speech}

\begin{table*}[t]
\centering
\small
\setlength{\tabcolsep}{2.5pt}
\begin{tabular}{l|cccc|cccc|c}
\toprule
\multirow{2}{*}{Method} & \multicolumn{4}{c|}{EER} & \multicolumn{4}{c|}{WER}
& \multirow{2}{*}{ALL AVG} \\
& meeting & banking & medical & AVG & conversation & meeting & dictation & AVG & \\
\midrule
Dense attention        & 28.43 & 8.72 & 7.25 & 14.80 & 21.53 & 19.61 & 11.42 & 17.52 & 16.16 \\
MoBA (dense infer)  & 26.23 & 12.66 & 8.40 & 15.76 & 26.69 & 19.06 & 10.97 & 18.91 & 17.34 \\
\textbf{LazyFlash}     & \textbf{23.04} & \textbf{8.66} & \textbf{6.61} & \textbf{12.77}
& \textbf{21.35} & \textbf{19.01} & \textbf{10.71} & \textbf{17.02} & \textbf{14.90} \\
\bottomrule
\end{tabular}
\caption{Long-form English ASR results on 10-minute utterances (lower is better,
$\downarrow$). MoBA is shown with dense inference only, the configuration under which it
performed best on text in Table~\ref{tab:text}.}
\label{tab:speech}
\end{table*}

Long-form ASR is a demanding test for a change to attention, because a dropped phrase or a hallucinated word is counted directly and the model cannot paraphrase its way around a passage it has failed to attend to. Besides the word error rate (WER), this paper reports the entity error rate (EER)~\citep{deng2026unique}, defined as one minus entity recall. A reference entity counts as correct only when every one of its words is transcribed verbatim, which makes EER a stricter measure of whether the key information in an information-dense recording has been captured.



Table~\ref{tab:speech} reports both metrics on three test sets each. LazyFlash
is ahead of dense attention on every column. The macro-average falls from 16.16
to 14.90 and the EER average from 14.80 to 12.77, with the largest gain on the
online-meeting EER, from 28.43 to 23.04. 
The gain may come from the greater redundancy of long-form speech, where 
neighboring acoustic frames are highly correlated and many query blocks 
contribute little new information. Skipping
low-gradient blocks may therefore reduce sensitivity to repetitive or noisy
acoustic context while retaining the blocks that matter to the update. MoBA is
reported with dense inference, the configuration under which it was strongest
on text, and still falls behind dense attention on the macro-average, 17.34
against 16.16.

The attention timings follow the pattern seen on text. LazyFlash spends 446.4~ms in the backward pass against 1091.9~ms for dense attention, a speedup of $2.45\times$, and its forward cost is unchanged at 391.7~ms against 389.4~ms. MoBA spends 613.7~ms in the forward pass, where its routing runs, and 704.3~ms in the backward.

\subsection{Training Efficiency}
\label{sec:efficiency}

\begin{figure*}[h]
\centering
\includegraphics[width=\textwidth]{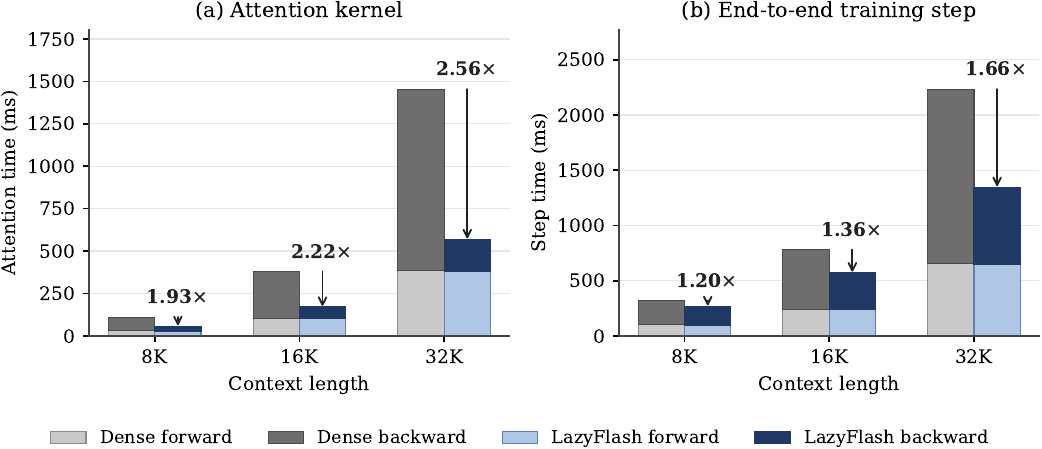}
\vspace{-0.7cm}
\caption{Attention time and end-to-end training step time against context length, measured
on real LongAlign examples truncated to each length. Forward and backward are shown
separately, and the arrows give the speedup over dense attention.}
\label{fig:scale}
\end{figure*}

LazyFlash removes work only from the attention backward, whose share of a training step grows with context length because attention scales quadratically with the sequence while the rest of the model scales linearly. Figure~\ref{fig:scale} measures both at 8K, 16K and 32K tokens.


\textbf{Attention kernel speedup.} At every length the forward pass matches dense attention,
and the whole saving comes from the backward pass, which speeds up by $2.77\times$ at 8K,
$3.97\times$ at 16K and $5.72\times$ at 32K. Two effects contribute. The attention backward
grows quadratically with length, so it takes a larger share of the attention cost. At the
same budget $\delta$, the measured share of computed blocks falls from $18\%$ at 8K to
$10\%$ at 32K, providing an additional source of speedup. Over forward and backward together, attention is $1.93\times$ faster at 8K
and $2.56\times$ at 32K, with the forward pass setting a floor on this ratio.

\textbf{End-to-end speedup.} At the level of a full training step the speedup is
$1.20\times$ at 8K, $1.36\times$ at 16K and $1.66\times$ at 32K, with the model backward
alone reaching $2.26\times$ at 32K. The longer the context, the larger the share of training
that LazyFlash removes.

\subsection{Ablations}
\label{sec:ablation}

\begin{table*}[t]
\centering
\small
\setlength{\tabcolsep}{4pt}
\begin{tabular}{l|c|cc|cccc|c}
\toprule
\multirow{2}{*}{Selection rule} & \multirow{2}{*}{Overall} & \multicolumn{2}{c|}{Language}
& \multicolumn{4}{c|}{Difficulty} & Attn bwd \\
& & En & Zh & Extreme & Hard & Moderate & Easy & (ms) \\
\midrule
Dense attention (no selection)   & 29.64 & 30.31 & 28.63 & 25.18 & 25.40 & 19.54 & 41.78 & 1429.8 \\
\midrule
Random                           & 21.18 & 21.56 & 20.81 & 19.84 & 16.03 & 13.87 & 29.89 & 523.8 \\
Top-$k$, attention mass          & 27.07 & 27.96 & 26.19 & 23.02 & 24.62 & 17.50 & 37.98 & 548.0 \\
Top-$k$, gradient norm           & 28.45 & 29.84 & 27.07 & 23.87 & 25.20 & 19.16 & 40.16 & 508.0 \\
\textbf{LazyFlash}               & \textbf{30.24} & \textbf{30.37} & \textbf{29.03} & \textbf{25.22}
& \textbf{27.51} & \textbf{20.04} & \textbf{40.88} & 507.4 \\
\bottomrule
\end{tabular}
\caption{Selection rules compared on LongBench-Pro. \emph{Random} drops query blocks with no
signal, the two \emph{top-$k$} rules keep a fixed fraction ranked by forward attention mass
and by output-gradient norm, and \emph{LazyFlash} ranks by the gradient norm but sets the
amount skipped with a budget on gradient mass. The
three alternatives share the CUDA implementation of LazyFlash and are tuned to match its backward
time; dense attention is a quality reference and is not matched in cost.}
\label{tab:ablation}
\end{table*}

\textbf{Selection rule.} Three alternative rules isolate what makes the selection work. \emph{Random} drops query
blocks at random, so it keeps the same number of blocks as LazyFlash but pays no attention
to which ones. \emph{Top-$k$ on attention mass} keeps the blocks that receive the most
attention in the forward pass, the signal that sparse forward attention uses, rather than
the gradient that LazyFlash uses. 
\emph{Top-$k$ on the gradient norm} ranks blocks by the same signal as LazyFlash but keeps a
fixed fraction of them, while the budget of LazyFlash lets the amount skipped follow the
gradient distribution of each head and each sequence. This rule is
meProp~\citep{sun2017meprop} carried over from linear layers to attention blocks.
All
three share the CUDA implementation of LazyFlash and their sparsity is tuned so that the attention
backward time is close to $507.4$~ms, so that a difference in quality is not bought with a
difference in compute.

As shown in Table~\ref{tab:ablation}, Random reaches 21.18, only $1.5$ points above the model before fine-tuning in Table~\ref{tab:text} and $8.5$ points below dense training, so skipping this much of the
backward without choosing what to skip undoes most of the fine-tuning. Ranking by attention
mass recovers much of that gap at 27.07 but stays $3.2$ points behind LazyFlash, even with
$8\%$ more backward time. A block can receive a large share of the attention and still
contribute almost nothing to the gradient, so attention mass is not the right quantity to
rank by. Ranking by the gradient norm is the strongest of the three at 28.45, at the same
backward time as LazyFlash ($508.0$~ms against $507.4$~ms), but still $1.8$ points behind.
The two use the same signal and skip the same amount of work. A fixed fraction has to be
chosen once for a whole run, while the budget $\delta$ in LazyFlash adapts to each head and
each sequence, keeping $18\%$ of the blocks at 8K and $10\%$ at 32K
(Section~\ref{sec:efficiency}). Top-$k$ also drops the same blocks at every step, whereas
those dropped by LazyFlash change. LazyFlash is the only rule in the table that matches
dense training.

\begin{table*}[t]
\centering
\small
\setlength{\tabcolsep}{2.5pt}
\begin{tabular}{l|cccc|cccc|c}
\toprule
\multirow{2}{*}{Setting} & \multicolumn{4}{c|}{EER} & \multicolumn{4}{c|}{WER}
& \multirow{2}{*}{ALL AVG} \\
& meeting & banking & medical & AVG & conversation & meeting & dictation & AVG & \\
\midrule
Dense attention & 28.43 & 8.72 & 7.25 & 14.80 & 21.53 & 19.61 & 11.42 & 17.52 & 16.16 \\
\midrule
 LazyFlash ($\delta=0.1$) & 28.43 & \textbf{8.22} & 7.40 & 14.68 & \textbf{20.48} & 19.71 & 11.05 & 17.08 & 15.88 \\
 LazyFlash ($\delta=0.3$) & 26.96 & 8.94 & 6.76 & 14.22 & 21.49 & 19.22 & 10.73 & 17.15 & 15.69 \\
 LazyFlash ($\delta=0.4$) & \textbf{23.04} & 8.66 & \textbf{6.61} & \textbf{12.77} & 21.35 & \textbf{19.01}
& \textbf{10.71} & \textbf{17.02} & \textbf{14.90} \\
\bottomrule
\end{tabular}
\vspace{-0.1cm}
\caption{Effect of the budget $\delta$ on long-form ASR (lower is better, $\downarrow$).
Training diverges at $\delta=0.5$.}
\label{tab:delta}
\end{table*}

\textbf{Sensitivity to the budget.} The budget $\delta$ controls how much of the backward pass is skipped. A larger value saves
more work but computes the gradient from fewer blocks, and how far it can be pushed depends
on the workload. On text, quality is best
at $\delta=0.1$ and degrades beyond it, falling to 22.32 at $\delta=0.3$. Long-form speech
carries far more redundancy and tolerates a much larger budget: Table~\ref{tab:delta} shows
that quality keeps improving up to $\delta=0.4$, the value used in
Section~\ref{sec:speech}, while $\delta=0.5$ destabilises training. 
At $\delta=0.1$, LazyFlash matches or improves on dense attention in both
modalities, making it a robust default.

\subsection{Model Size Scaling}
\label{sec:scaling}

\begin{table*}[h]
\centering
\small
\setlength{\tabcolsep}{4pt}
\begin{tabular}{lc|cc|cccc|cc}
\toprule
\multirow{2}{*}{Method} & \multirow{2}{*}{Overall} & \multicolumn{2}{c|}{Language}
& \multicolumn{4}{c|}{Difficulty} & \multicolumn{2}{c}{Training attn.\ (ms)} \\
& & En & Zh & Extreme & Hard & Moderate & Easy & fwd & bwd \\
\midrule
\multicolumn{10}{l}{\emph{Qwen3-4B-Base}} \\
\quad before fine-tuning & 6.88 & 5.84 & 7.92 & 11.09 & 4.91 & 4.08 & 5.89 & -- & -- \\
\quad Dense attention    & 24.79 & 23.33 & \textbf{26.26} & 22.32 & 18.56 & 15.92 & 36.11 & 534.4 & 1427.0 \\
\quad \textbf{LazyFlash} & \textbf{26.55} & \textbf{27.18} & 25.92 & \textbf{22.94} & \textbf{23.33} & \textbf{18.34} & \textbf{36.72} & 544.0 & \textbf{500.7} \\
\midrule
\multicolumn{10}{l}{\emph{Qwen3-8B-Base}} \\
\quad before fine-tuning & 19.70 & 19.35 & 19.78 & 19.39 & 16.89 & 13.98 & 24.67 & -- & -- \\
\quad Dense attention    & 29.64 & 30.31 & 28.63 & 25.18 & 25.40 & 19.54 & \textbf{41.78} & 547.3 & 1429.8 \\
\quad \textbf{LazyFlash} & \textbf{30.24} & \textbf{30.37} & \textbf{29.03} & \textbf{25.22} & \textbf{27.51} & \textbf{20.04} & 40.88 & 562.2 & \textbf{507.4} \\
\midrule
\multicolumn{10}{l}{\emph{Qwen3-14B-Base}} \\
\quad before fine-tuning & 26.37 & 25.81 & 26.94 & 23.68 & 24.48 & 18.71 & 34.57 & -- & -- \\
\quad Dense attention    & 31.17 & 31.05 & 31.29 & 26.42 & 28.59 & 22.38 & 42.33 & 759.7 & 1917.6 \\
\quad \textbf{LazyFlash} & \textbf{33.60} & \textbf{33.73} & \textbf{33.46} &
\textbf{27.60} & \textbf{30.58} & \textbf{24.65} & \textbf{46.24} & 779.5 &
\textbf{654.2} \\
\bottomrule
\end{tabular}
\caption{LongBench-Pro accuracy and per-step attention time across backbone
sizes. Qwen3-4B and Qwen3-8B have the same attention configuration, and their
dense attention times are nearly identical. LazyFlash wins 19 of the 21
accuracy splits.}
\label{tab:backbones}
\end{table*}


Building on the main experiment of Section~\ref{sec:text}, the same setup is
applied to Qwen3-4B and Qwen3-14B to verify that LazyFlash remains effective at
different model sizes. The data mixture, the number of steps, the global batch
size and the budget $\delta=0.1$ are unchanged. The dense baseline improves with
size, from 24.79 to 29.64 to 31.17 overall, while LazyFlash remains ahead by
$+1.76$ at 4B, $+0.60$ at 8B and $+2.43$ at 14B, winning 19 of the 21 splits
and all seven at 14B. The attention backward is $2.85\times$, $2.82\times$ and
$2.93\times$ faster at 4B, 8B and 14B, respectively, while the forward time
remains close to dense attention. These results show that LazyFlash consistently
accelerates long-context fine-tuning across model sizes while maintaining
accuracy.

\section{Conclusions}

This paper proposes LazyFlash, a forward-dense and backward-sparse attention that
accelerates long-context fine-tuning without making the trained model sparse.
LazyFlash scores each query block by its output-gradient norm, sets how much of the
backward may be skipped with a budget on gradient mass, and draws the blocks it keeps
by unbiased sampling; the CUDA selection pipeline produces the per-block weights and the
FlashAttention backward consumes them, so a skipped block is never loaded. On
LongBench-Pro across three Qwen3 backbones and on long-form ASR, LazyFlash matches
dense attention in both text and speech while cutting attention backward time by
$2.82\times$ and $2.45\times$. In the scaling benchmark it reaches up to
$5.72\times$ attention-backward and $1.66\times$ end-to-end model training speedup at 32K
context, and because the forward is exact the trained model remains an ordinary dense
model that any standard inference stack can serve.

\newpage

\subsection*{AI use statement}

In this work, we used generative AI tools for language polishing and grammar correction. We have not used generative AI tools for research ideation, data collection, model design, or result analysis, and other generative AI-assisted tasks are not applicable to this work. All AI-assisted edits were carefully reviewed and verified by the authors to ensure that the original academic meaning, technical claims, and factual accuracy were strictly preserved without introducing hallucinated content or altered claims. We take full responsibility for the final content of this work, including all text, claims, and artifacts.

\subsection*{Reproducibility statement}


Section~\ref{sec:method} specifies the block score, gradient-mass budget and
sampling distribution, while Appendix~\ref{app:repro} provides pseudocode and
kernel details. Appendix~\ref{app:hyper} lists the complete training
configuration. The text experiments use public Qwen3 checkpoints, LongAlign-10k,
ShareGPT and LongBench-Pro, allowing them to be reproduced from public artifacts.

\bibliography{iclr2027_conference}
\bibliographystyle{iclr2027_conference}

\appendix
\section{Hyper-parameters}
\label{app:hyper}

Across all experiments LazyFlash operates on query-row blocks of $128$ tokens
against key blocks of $128$ tokens. The
forward pass is bit-exact dense in every run, so the budget $\delta$ is the only
method hyper-parameter.

For text fine-tuning, Qwen3-4B-Base, Qwen3-8B-Base and Qwen3-14B-Base are fully
fine-tuned on a $1{:}1$ mixture of LongAlign-10k and ShareGPT, with the loss
placed on the final assistant turn only. Training runs for $2500$ steps at a
maximum sequence length of $65\,536$ on eight H200 GPUs. All three model sizes
use two sequences per device and two gradient accumulation steps, giving a
global batch of $32$ sequences. The
optimizer is AdamW ($\beta_1{=}0.9$, $\beta_2{=}0.95$, $\epsilon{=}10^{-8}$, no
weight decay) at a learning rate of $10^{-5}$ with $60$ linear warm-up steps
followed by linear decay, gradient clipping at $1.0$, bf16 mixed precision, and
DeepSpeed ZeRO-$2$. YaRN scaling extends the
$32\,768$-token context of the released checkpoints by a factor of four.
LazyFlash uses $\delta=0.1$. The MoBA baseline uses a block size of $512$ and
$k{=}4$ under the convention of the original paper, where the current block
counts toward $k$, giving a budget of $2048$ tokens per query.

For speech, a long-form ASR model is built by attaching a $24$-block Conformer
encoder (attention dimension $1024$, $16$ heads, FFN $1536$, kernel size $3$,
$8\times$ time reduction, T5 relative position bias with maximum distance $500$)
to Qwen3-8B. The language backbone, its final norm, the embeddings and the
output head are frozen; the encoder, the adapter and LoRA (rank $r{=}320$,
$\alpha{=}640$, dropout $0.01$) on all attention and MLP linear layers are
trained. Training runs for $10\,000$ steps on English long-form data with
utterances between $60$ and $1800$ seconds, packed
into $32\,768$-token batches with one batch per device and a global batch of
$16$. The learning rate is $4\!\times\!10^{-5}$ with $100$ linear warm-up steps,
AdamW uses $\beta_1{=}0.9$, $\beta_2{=}0.95$, $\epsilon{=}10^{-7}$ and weight
decay $0.01$, with gradient clipping at $1.0$, DeepSpeed ZeRO-$1$, and bf16 mixed
precision. Since the loss is text-only and audio
positions are ignored, the output gradient is far more concentrated than in
text, and LazyFlash uses the larger budget $\delta=0.4$.

\section{Implementation and Reproducibility Details}
\label{app:repro}

Algorithm~\ref{alg:select} states the per-$(b,h)$ selection that runs once per
backward call, where $M$ is the number of query-row blocks and $c_m$ the
Frobenius norm of the output gradient of block $m$.

\begin{algorithm}[h]
\caption{LazyFlash block selection for one $(\text{batch},\text{head})$}
\label{alg:select}
\begin{algorithmic}[1]
\REQUIRE $dO \in \mathbb{R}^{S \times d}$, budget $\delta$
\STATE $c_m \gets \| dO_{[128m\,:\,128(m+1)]} \|_F$ \quad for $m = 1 \dots M$
\STATE sort $c^2$ ascending and let $k^\ast$ be the largest $k$ with
       $\sum_{i\le k} c^2_{(i)} \le \delta \sum_m c_m^2$
\STATE find $\lambda$ by bisection such that $\sum_m \min(1, \lambda c_m) = M - k^\ast$
\STATE $q_m \gets \min(1, \lambda c_m)$
\STATE draw $u_m \sim \mathcal{U}(0,1)$ and set $w_m \gets 1/q_m$ if $u_m < q_m$, else $w_m \gets 0$
\STATE $W[m,n] \gets w_m$ for all causally valid key blocks $n$
\RETURN $W \in \mathbb{R}^{M \times M}$
\end{algorithmic}
\end{algorithm}

Blocks with large $c_m$ saturate at $q_m = 1$, so they are never dropped and
never rescaled; only the low-gradient tail is sampled. A kept block carries the
weight $w_m = 1/q_m$ and a skipped block carries $w_m = 0$, so
$\mathbb{E}[w_m] = q_m \cdot (1/q_m) = 1$ for every block and the resulting
gradient estimate is unbiased.

The implementation is based on FlashAttention $2.5.2$. The selection above uses
three CUDA kernels: one reduces $dO$ into per-block energies, one applies the
budget and the bisection per $(\text{batch},\text{head})$, and one broadcasts the
block weight across key blocks, producing an fp32 weight tensor of shape
$[B, H, M, N]$. The backward kernel reads this tensor and iterates only over
active query blocks, prefetching the next active block's $Q$ and $dO$ so that the
software pipeline of the dense kernel is preserved; a skipped block costs a
single weight read and no tile traffic. The weight $1/q_m$ is applied inside the
kernel, so no rescaling pass is needed.

Attention timings are collected with CUDA events bracketing the interval from the
attention output to the query gradient through autograd hooks, which isolates the
attention backward from the rest of the layer. Dense attention, LazyFlash and
MoBA are run from the same container image and the same compiled kernel, so the
attention path is the only difference between them. All measurements reported in
this paper were taken on H200 GPUs.

\end{document}